\documentclass{article}

\usepackage{PRIMEarxiv}

\usepackage[utf8]{inputenc} 
\usepackage[T1]{fontenc}    
\usepackage{hyperref}       
\usepackage{url}            
\usepackage{booktabs}       
\usepackage{amsfonts}       
\usepackage{nicefrac}       
\usepackage{microtype}      
\usepackage{lipsum}
\usepackage{fancyhdr}       
\usepackage{graphicx}       
\graphicspath{{media/}}     

\usepackage{graphicx}
\usepackage[numbers]{natbib}
\usepackage{booktabs}
\usepackage[utf8]{inputenc}
\usepackage{CJKutf8}
\usepackage{soul}
\usepackage{ulem}
\usepackage[british]{babel}
\usepackage[T1]{fontenc}

\usepackage[norelsize, linesnumbered, ruled, lined, boxed, commentsnumbered]{algorithm2e}

\usepackage{algpseudocode}

\newcommand{\ie}{\textit{i}.\textit{e}., }

\title{Transformer-based Approaches \\for Legal Text Processing
\thanks{\textit{\underline{Citation}}: 
\textbf{Ha-Thanh Nguyen et al. Transformer-based Approaches for Legal Text Processing. Pages: 1-21 DOI: 10.1007/s12626-022-00102-2}} 
}

\author{
  Ha-Thanh Nguyen, Phuong Minh Nguyen, Quan Minh Bui \\
   Japan Advanced Institute of Science and Technology \\
  \And
  Chau Minh Nguyen, Binh Tran Dang, Minh Le Nguyen\\
  Japan Advanced Institute of Science and Technology \\
   \And
  Vu Tran \\
  Institute of Statistical Mathematics\\
  \And
  Thi-Hai-Yen Vuong \\
  University of Engineering and Technology, VNU\\
  \And
  Ken Satoh \\
  National Institute of Informatics\\
}

\begin{document}
\maketitle

\begin{abstract}
In this paper, we introduce our approaches using Transformer-based models for different problems of the COLIEE 2021 automatic legal text processing competition.
Automated processing of legal documents is a challenging task because of the characteristics of legal documents as well as the limitation of the amount of data.
With our detailed experiments, we found that Transformer-based pretrained language models can perform well with automated legal text processing problems with appropriate approaches.
We describe in detail the processing steps for each task such as problem formulation, data processing and augmentation, pretraining, finetuning.
In addition, we introduce to the community two pretrained models that take advantage of parallel translations in legal domain, NFSP and NMSP.
In which, NFSP achieves the state-of-the-art result in Task 5 of the competition.
Although the paper focuses on technical reporting, the novelty of its approaches can also be an useful reference in automated legal document processing using Transformer-based models.
\end{abstract}


\section{Introduction}
Case law and statute law are the two largest sources of law in the world, based on these two sources of law, social relations are adjusted in accordance with international practices and national laws. 
For case law, the cases that are heard first will be used as the basis for handling the following cases.
In statute law, legal documents are the main basis for court decisions.
The competition on building automated legal text processing systems is a challenging and inspiring one.
In COLIEE 2021, there are three kinds of tasks in automated models include retrieval, entailment, and question answering.
Taking advantage of Transformer-based models, we achieve competitive performance in this competition.
In this paper, we report in detail our approaches and analysis the results in using the Transformer-based models in dealing with legal text processing tasks.

In order to serve the readability of the article, we provide the information about 5 tasks of COLIEE 2021.
Task 1 is a case law retrieval problem. With a given case law, the model needs to extract the cases that support it. This is an important problem in practice. It is actually used in the attorney's litigation as well as the court's decision-making.
Task 2 also uses caselaw data, though, the models need to find the paragraphs in the existing cases that entail the decision of a given case.
Task 3, 4, 5 uses statute law data with challenges of retrieval, entailment, and question answering, respectively.
From a design perspective, these 5 tasks cover the main tasks in automatic legal document processing.

One of the main challenges of legal domains in general and COLIEE 2021, in particular, is the scarcity and difficulty of data.
For the traditional deep learning approach, the amount of data provided by the organizer is difficult for constructing effective models. For that reason, we use pretrained models from problems that have much more data and then finetune them for the current task.
In Tasks 1, 2, 3, and 4 we have used lexical score and semantic score to filter a correct candidate. In our experiments, the ratio is not 50:50 for lexical score and semantic score, we find out that the increase in the rate of lexical score leads to efficiency in ranking candidates. 
In task 5, the system relies on a pretrained model with multilingual data to make predictions.
We have achieved state-of-the-art performance in the blind test set provided by the organizer for this task.

The challenge we always face in real-world competitions is that the amount of labelled data is often quite modest.
One of the features of our approaches is data engineering. Based on labelled standard data (\ie gold data), we generate augmented samples (\ie silver data) in massive quantities, at the expense of the existence of noise points in the data.
Based on experiments and observations in model construction, we choose the most suitable configuration for each specific task.

Transformer \cite{vaswani2017attention} architecture and its variants like BERT~\cite{devlin2018bert}, ALBERT~\cite{lan2019albert}, ELECTRA~\cite{clark2020electra}, BART \cite{lewis2019bart}, GPTs~\cite{radford2018improving, radford2019language, brown2020language} have made many breakthroughs in the field of natural language processing.
Through problem analysis, we propose solutions using deep learning with Transformer-based models. 

\begin{figure}
  \centering
  \includegraphics[width=.6\columnwidth]{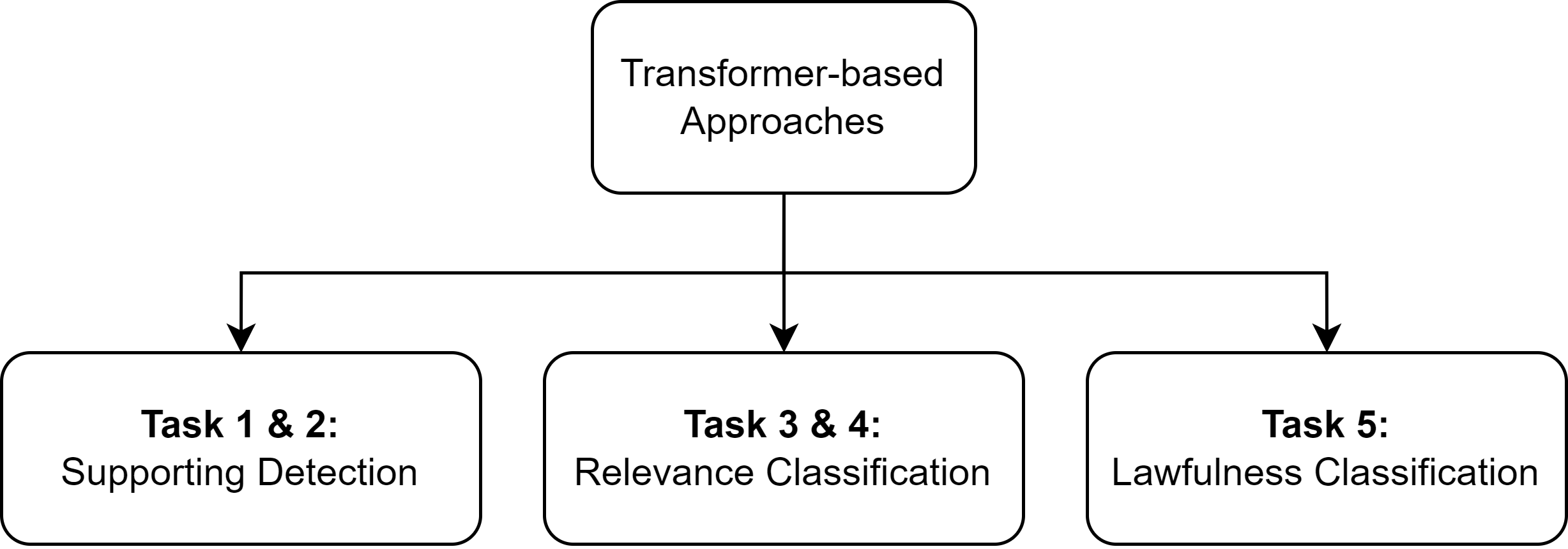}
  
  \caption{Overall view point of our Transformer-base approaches.}
  \label{fig:overall}
\end{figure}

Figure \ref{fig:overall} demonstrates our Transformer-based approaches for each COLIEE task.
With the multi-headed self-attention mechanism and the huge pretraining data amount, these models can deal with the data scarcity.
Dealing with Task 1 and Task 2, we introduce an approach namely supporting detection.
Solving Task 3 and Task 4, we propose the approach of relevance classification with other supplement techniques.
In Task 5, the idea of using multilingual resources introduced in NMSP and NFSP and other techniques in data enrichment is also the novelty of this paper.
Our proposals of deep learning methods can be a useful reference for researchers and engineers in automated legal document processing.




\section{Related Works}
\subsection{Legal Text Processing}
Processing legal documents is always challenging and there are many different approaches to different problems in this field.
Legal data consists of long sentences, contains many specialized terms, contains complex logical meanings, and often requires a high level of understanding.
Therefore, legal text processing is a difficult class of problems in NLP to which for all approaches are difficult to provide a complete solution.

The first approach is to treat legal documents as collections of logical constraints \cite{kowalski2021logical}.
Logical relationships are an important feature in the semantics of legal documents. Any regulation of social relations by law must be based on logic. Hence, here is a sound approach. From the application aspect, PROLEG \cite{satoh2010proleg} is one of the famous systems that implement logical inference in the legal domain.
This system can perform logical inferences to make legal decisions based on the provided rules and facts.

\begin{figure*}
    \centering
    \includegraphics[width=0.5\textwidth]{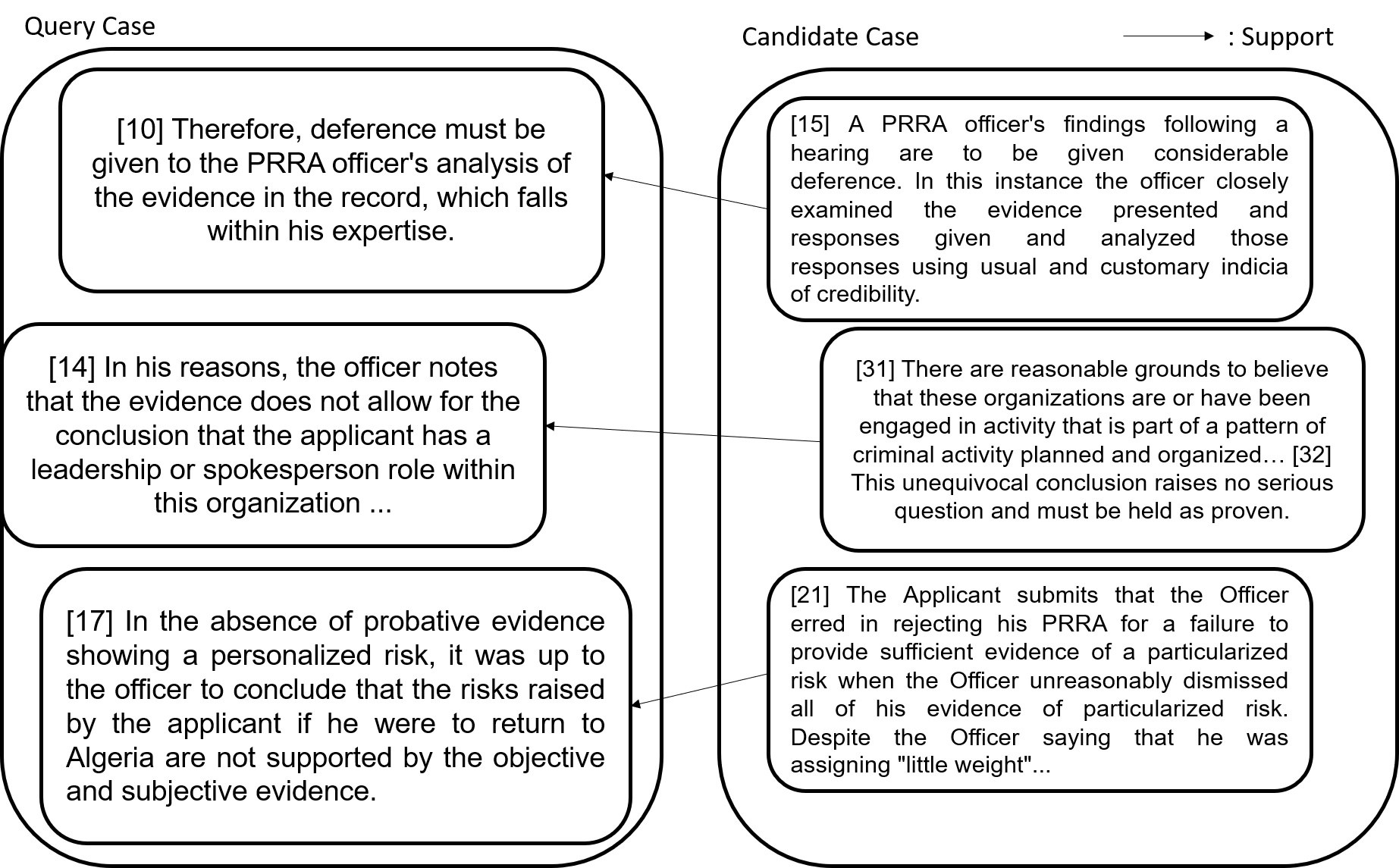}
    \caption{An example of supporting relationship among legal contents.}
    \label{supporting_definition}
\end{figure*}

The second approach is to use morphological features in processing legal documents.
Legal documents often have specialized terms. These words contain large amounts of information when they are used.
In addition, these terms often convey clear semantics in their expressions.
So systems (especially retrieval systems) based on lexical features can achieve acceptable results in the legal domain.
These systems use fairly classical NLP techniques and theories \cite{cooper1971definition,luhn1957statistical,salton1988term}. At the expense of simplicity, these systems often have problems with tasks that require natural language understanding.

A third approach is to use machine learning to statistically compute features from the data and make predictions. In this approach, artificial neural networks \cite{sugathadasa2018legal,kien-etal-2020-answering} and especially transformer-based models \cite{devlin2018bert,liu2019roberta,lewis2019bart}, are techniques that have made many achievements in recent years. These methods, although powerful, have the disadvantage of requiring a large amount of training data and comprehension of the data characteristics in the downstream tasks.
This has also been a strong approach at COLIEE in recent years.

\subsection{Case Law Processing at COLIEE} 
In COLIEE 2018, most teams chose the lexical-based approaches. UBIRLED ranked the candidate cases based on tf-idf. They got approximately 25\% of the candidate cases with the highest scores. UA and several teams also chose the same approach with UBIRLED. They compared lexical features between a given base case and corresponding candidate cases. JNLP team combined lexical matching and deep learning, which achieved state-of-the-art performance on Task 1 with the F1 score of 0.6545.

In COLIEE 2019, several teams applied machine learning including deep learning to both tasks. JNLP team achieved the best result of Task 1 in COLIEE 2019 \cite{tran2019building} using an approach similar to theirs in COLIEE 2018. In Task 2, their deep learning approach achieved lower performance compared to their lexical approach. Team UA's combination of lexical similarity and BERT model achieved the best performance for Task 2 in COLIEE 2019~\cite{10.1145/3322640.3326741}.

In COLIEE 2020, the Transformer model and its modified versions were widely used. TLIR and JNLP~\cite{nguyen2020jnlp} teams used them to classify candidate cases with two labels (support/non-support) in Task 1. Team cyber encoded candidate cases and the base case in tf-idf space and used SVM to classify. They demonstrated the ability of the approach with the first rank in Task~1. In Task 2, JNLP~\cite{nguyen2020jnlp} continually applied the same approach in Task 1 in the weakly-labeled dataset. It made them surpass team cyber and won Task~2.

In COLIEE 2021, the teams applied the combination of linguistic information, machine learning methods and pretrain models to solve the problem. The UA team used BERT and Naive Bayes classifier. The NM team combined sequence-to-sequence models such as monoT5-zero-shot, monoT5 and DeBERTa. JNLP team continues to use supporting models to generate weak labels for training. Other teams used BM25 and BERT to choose correct cases. 

\subsection{Statute Law Processing at COLIEE}
The retrieval task (Task 3) is often considered as a ranking problem with similarity features.
In COLIEE 2019, most of the teams used lexical methods for calculating the relevant scores. JNLP~\cite{jnlp_3_2019}, DBSE~\cite{dbse_3_2019} and IITP~\cite{iitp_3_2019} chose tf-idf and BM25 to build their models. JNLP used tf-idf of noun phrase and verb phrase as keywords which show the meaning of statements in the cosine-similarity equation. DBSE applied BM25 and Word2vec to encode statements and articles. The document embedding was used to calculate and rank the similarity score. Team KIS~\cite{kis_3_2019} represented the article and query as a vector by generating a document embedding. Keywords were selected by tf-idf and assigned with high weights in the embedding process. In COLIEE 2020, with the popularity of Transformer based methods, the participants change their approaches. The task winner, LLNTU, only used BERT model to classify articles as relevant or not.
    
Regarding the entailment task (Task 4), approaches using deep learning have attracted more attention. In COLIEE 2019, KIS~\cite{kis_4_2019} used predicate-argument structure to evaluate similarity. IITP~\cite{iitp_3_2019} and TR~\cite{tr_4_2019} applied BERT for this task. JNLP~\cite{jnlp_task4_coliee2019} classified each query to follow binary classification based on big data. In COLIEE 2020, BERT and multiple modified versions of BERT were used. JNLP~\cite{nguyen2020jnlp} chose a pretrained BERT model on a large legal corpus to predict the correctness of statements.

In COLIEE 2021, the teams continued using BERT as main model of their approaches. The HUKB used BERT and data augmentation. JNLP team applied bert-base-japanese-whole-word masking and ensemble models (original multilingual BERT, Next Foreign Sentence Prediction pretrain model and Neighbor Multilingual Sentence Prediction pretrain model). The UA team built an information retrieval system based on methods such as BM25 (UA.BM25), TF-IDF(UA.tfidf) and language model (UA.LM).

\section{Method}
\label{sec:method}

\subsection{Task 1 and Task 2. Case Law Processing}
In COLIEE 2021, there is a significant change in the data structure in Task 1, which brings more challenges for all participated teams. 
This can be seen in the official results of the organizers, which we will quote later in this Section.
The performance of participants' models drops badly from nearly 70\% to 20\% or worse
The number of candidate legal cases for each query increases from 200 in last year's competition to 4,415 in this year's competition.
With a naive estimate, the difficulty of this year's problem is at least 22 times higher than last year's problem.
The larger the search space, the lower the probability that the model correctly predicts a relevant case.

\begin{figure*}
    \centering
    \includegraphics[width=0.5\textwidth]{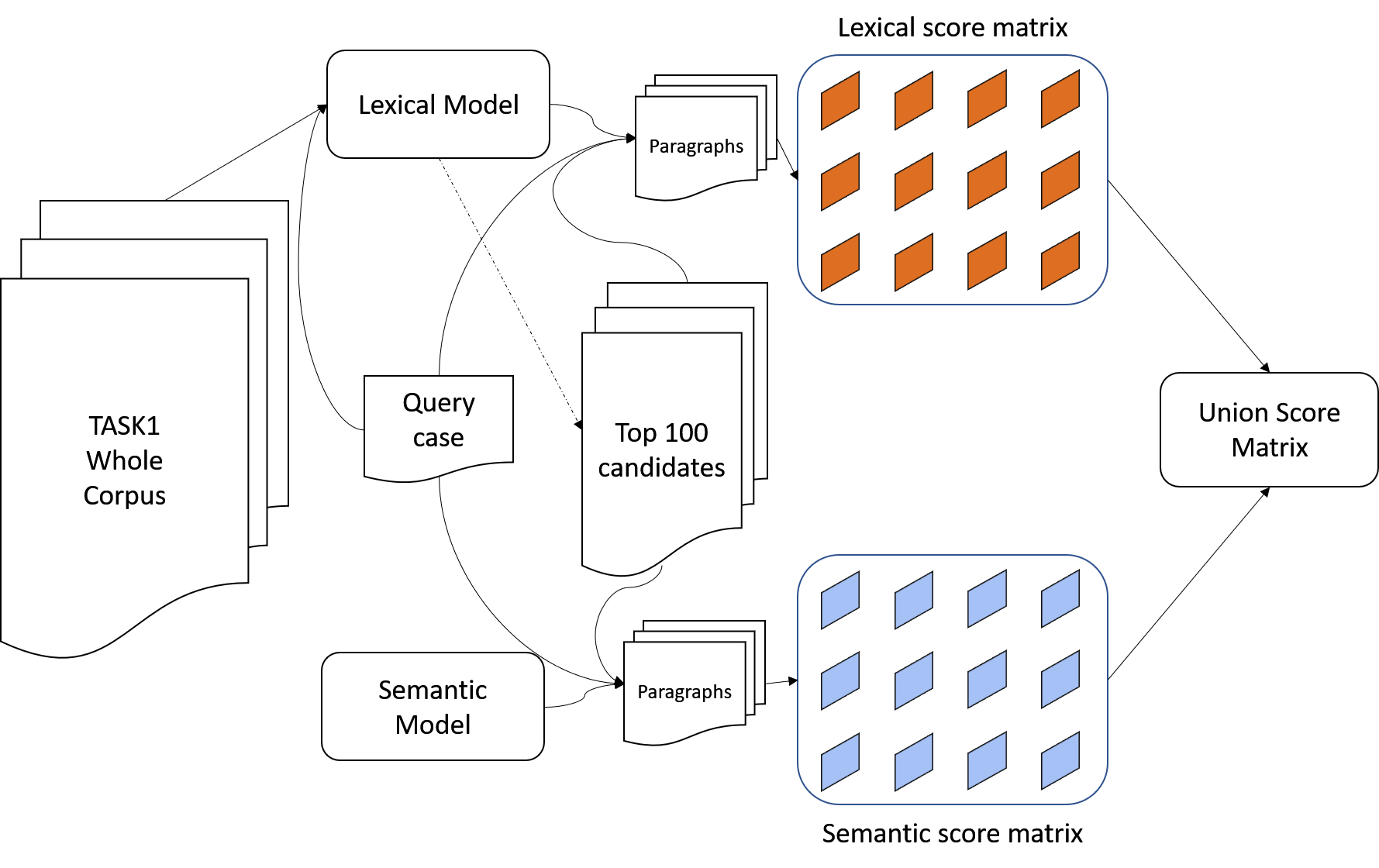}
    \caption{Demonstration of mixing lexical and semantic score.}
    \label{Union_score}
\end{figure*}

Figure \ref{Union_score} demonstrates the flow of our approach.
4,415 candidates are too large for a deep learning model to work effectively and efficiently. 
Therefore, we use BM25 to filter out 100 cases that lexically match the query before processing with the deep learning model.
In both Task 1 and Task 2, we evaluate candidates based on the Union Score, a metric that combines the lexical score and the semantic score.
Through experiments, we confirm that this approach is more efficient than using only lexical or semantic features.

The previous approaches focus on encoding the whole case law and calculating the similarity of vector representation for each query-candidate pair of cases. In our method, we handle the similarity between candidates on the paragraph level. ``Supporting'' and ``Relevant'' are subtractive definitions, and we assume that for each paragraph in the query case, there are one or more paragraphs that carry useful information for the given query's paragraph. As we can see in the Figure \ref{supporting_definition}, paragraph [10] has \textit{``PRRA officer's''} and in the candidate case number [15] also has \textit{``PRRA officer's''}, there seems to be a lexical relationship here. This is evidence that the lexical model could be effective. Along the line of the query paragraph number [14] has \textit{``the evidence''} corresponding to \textit{reasonable grounds} in the candidate paragraph number [31], we can see the semantic relationship between two paragraphs through this example. For discovering the relevant paragraph, we combine the lexical similarity and the semantic similarity score.

\textbf{Lexical Matching} For the purpose of capturing lexical information, we use Rank-BM25~\footnote{https://pypi.org/project/rank-bm25/}, a collection of algorithms for querying a set of documents and returning the ones most relevant to the query. For Task 1, after reducing the searching space to 100 queries, we separate the query case and also candidate cases into paragraphs. Assume that the query case has $N$ paragraphs, and the candidate case has $M$ paragraphs, we will calculate the lexical mapping score by Rank-BM25 for each query paragraph and every single paragraph in the candidate case. Then we obtain the matrix lexical score size $N \times M$ for each query-candidate pair. We keep these matrixes and their union with supporting score which is introduced later. 

\textbf{Supporting Matching} As we mentioned in the Figure \ref{supporting_definition}, we extract the semantic relationship between query paragraph and candidate paragraph. To obtain the relationship score, we use pretrained model BERT provided by huggingface\footnote{https://huggingface.co/models}. The same approach as lexical matching, we sequentially split a query and corresponding candidate cases into paragraphs, and obtain a matrix semantic score that has the same size as matrix lexical score. 
Although BERT is a reputable pretrained model that achieves state-of-the-art results on many NLP tasks, it still has limitations when applied to COLIEE data.
In particular, the law is specific data while BERT is trained on the general domain data.
For that reason, we develop a silver dataset based on the COLIEE dataset for finetuning BERT and use this for predicting the semantic relationship between paragraphs.

\textbf{Union Score} To obtain the most relevant cases for a given query, we find a semantic as well as lexical relationship overlapped between the two paragraphs. For the exact purpose of developing union score we use the following formula:
\begin{equation} \label{combine_score}
    union\_score = \alpha * score_{supporting} + (1-\alpha) * score_{BM25}
\end{equation}

In Task 2, we use the same method as Task 1 with a few improvements. 
We consider Task 2 as a binary classification problem with the training data as a set of sentence pairs. With this approach, we can obtain more gold training data to train the supporting model. For optimizing the performance of the supporting model on this problem, after finetuning on the silver data, we finetune one more time on the gold data of Task 2.

\subsection{Task 3. The Statute Law Retrieval Task\label{subs:3}} This task requires participants to determine whether a set of articles entail a legal bar exam question: Entails($A_{i  \mid 1 \leq i \leq n}$, $Q$) or Entails($A_{i  \mid 1 \leq i \leq n}$, \textit{not} $Q$), where $Q$ is a legal bar exam question, and $A_{i  \mid 1 \leq i \leq n} $ is a subset of articles in Japanese Civil Code. This task faces two main challenges. The first challenge is to answer questions about a specific legal case. It is notable that while articles in statute law are described in an abstract manner, specific legal case reports a specific event. This difference requires a model to have reasoning ability. The second challenge is to deal with the long articles. In this section, we focus on tackling the second challenge by employing two techniques: (i) text chunking technique on prepared training data, and (ii) self-labeled technique in the model finetuning phase.

\paragraph{\textbf{Training data preparation.}}
In the training data preparation phase, we employ the method proposed in \cite{nguyen2020jnlp} with a modification. Firstly, we obtain positive training examples by pairing a question with its annotated entailing articles. The negative training examples are the pair of a question with any other articles. Because of the characteristics of the evaluation dataset, the negative training examples are dominant. We then remove negative training examples based on the tf-idf scores so that the number of negative examples corresponding to a question is 150 at maximum. By using the top 150 Articles most related to the question via Tf-IDF score, the recall score can archieve to 91.25\%. The details can be found in our previous paper \cite{nguyen2020jnlp}.

\paragraph{\textbf{Text chunking technique.}}
In this task, we leverage the language understanding ability of Transformer-based pretrained language models. Because of the 512-token limitation in the design of those models, they have to truncate a long article if it contains more than 512 tokens. It leads to the lack of article information when we process the article. Furthermore, in fact, for a question, in most cases, there are only a few parts of articles that entail the question (Table~\ref{tab:example_chunk}). To overcome the 512-token limitation, we utilize a sliding window to obtain one or multiple chunks from an article. We experimented with many values of sliding window and stride to get the most appropriate values. Now, we consider ($question$, $chunk$) as training examples instead of the original ($question$, $article$), where the label for ($question$, $chunk$) is derived from the corresponding ($question$, $article$).

\paragraph{\textbf{Self-labeled technique.}}
We observe that there are many noises in the generated training data following the above process. For example, suppose that sub-articles $A_{i,1}, A_{i,2}, ..., A_{i,j}$ is obtained from a long article $A_{i}$ which entails a question $Q$, and only $A_{i,j}$  (the sub-article $j^{th}$ of article $A_i$) entails $Q$. In this case, our training data generation process will label all pairs ($Q, A_{i,1}$), ($Q, A_{i,2}$), ..., ($Q, A_{i,j}$) as positive training examples. However, only pair ($Q, A_{i,j}$) should be labeled positive, while other pairs, i.e. ($Q, A_{i,1}$), ($Q, A_{i,2}$), ..., ($Q, A_{i,j-1}$), should be labeled as negative examples. Inspired by the self-labeled techniques \cite{triguero2015self}, we propose to mitigate the number of incorrect generated positive training examples by employing a simple self-labeled technique. 
Let $\mathbf{X}=\{ x_1, x_2, ..., x_k\}$ be generated input training samples that are pairs of $Q$ and sub-articles where $k$ is the number of generating training samples, $\mathbf{y^0}$ be the initial labels of these samples drawn from the annotated label for pair of $Q$ and corresponding article, $\mathcal{M}$ denotes the model and also its forward function, the self-labeld technique is described in Algorithm \ref{alg:self-label}.


\begin{algorithm}
\SetAlgoLined
 \LinesNumbered
 \SetKwInOut{Input}{Input}
 \Input{Pairs of (Question, sub-Article) $\mathbf{X}$,  initial labels $\mathbf{y}^0$ } 
 \SetKwProg{Function}{function}{}{end}
 \SetKwRepeat{Do}{do}{while} 
 \Function{($\mathbf{X}$, $\mathbf{y^0}$)}{
    $\mathbf{y} = \mathbf{y}^0$; \\
        $\mathcal{M} \gets \textrm{fine-tuning} (\mathbf{X} ,  \mathbf{y} )$; // first fine-tuning to remove noise samples\\ 
        $\mathbf{\hat{y}} = \mathcal{M}(\mathbf{X})$;  \\ 
        $\mathbf{y}_i= \mathbf{\hat{y}}_i \forall i \in [1,k] : \mathbf{y}_i=True \land \mathbf{\hat{y}}_i \neq \mathbf{y}_i$;  // modifying the label \\
      $\mathcal{M} \gets \textrm{fine-tuning} (\mathbf{X} ,  \mathbf{y} )$;\\ 
        
 }
 \caption{Self-labeled technique to mitigate incorrect generated positive training examples.}\label{alg:self-label}
\end{algorithm}

In detail, firstly, the generated training examples are used to finetune a pretrained language model. After that, the finetuned model predicts labels for the generated training examples (where it is finetuned for). Next, labels of the generated training data is modified to mitigate noise samples. The modified labels then are utilized in the next self-labeled process. 
For the rule of modifying the label, base on our observation, we only allow the positive label to turn to a negative label, but not the reverse direction. Our experiments demonstrate the decrease in the incorrect positive training examples after employing the self-labeled technique.

\paragraph{\textbf{Model ensembling.}}
Since each finetuned language model has its advantages and disadvantages for different types of legal bar exam questions, we ensemble the prediction scores of the models with learned weights. It means that the prediction of a model may have a larger weight than the others on contributing to the final predictions.

\begin{table}[!htbp]
\centering
\begin{tabular}{lp{0.8\textwidth}}
\toprule
\textbf{Q. R01-4-E}          & In cases any party who will suffer any detriment as a result of the fulfillment of a condition intentionally prevents the fulfillment of such condition, the counterparty may deem that such condition has been fulfilled. \\\midrule
\textbf{Article 130}          &  Part I General Provisions Chapter V Juridical Acts Section 5 Conditions and Time Limits  \\ & (Prevention of Fulfillment of Conditions)          \\ & \textbf{(1) If a party that would suffer a detriment as a result of the fulfillment of a condition intentionally prevents the fulfillment of that condition, the counterparty may deem that the condition has been fulfilled.}
\\ & (2) If a party who would enjoy a benefit as a result of the fulfillment of a condition wrongfully has that condition fulfilled, the counterparty may deem that the condition has not been fulfilled. \\
\bottomrule
\end{tabular}
\caption{An example in which only one part of the entailing article entails the corresponding question. \label{tab:example_chunk}}
\end{table}

\subsection{Task 4. The Legal Textual Entailment Task}

This task involves the identification of an entailment relationship between relevant articles $A_{i  \mid 1 \leq i \leq n} $ (which is derived from Task 3's results) and a question $Q$. The models are required to determine whether the relevant articles entail ``$Q$'' or ``$not Q$''.  Given a pair of legal bar exam questions and article $(Q, A_i)$, the models return a binary value for determining whether $(A_i)$ entails $(Q)$. To address this task, we modified the training data preparation step, then use the same model architecture in Task 3 (Section~\ref{subs:3}) for training. 

\paragraph{Data preparation.} Based on our observation, the challenge of this task is to extract the relevance between a question and articles for classification while the number of given articles is relatively small (usually 1 or 2 articles are given). We hypothesize that the model can extract information more effectively and consistently if there are more relevant articles given. Therefore, we adapt the data augmentation technique mentioned in Task 3 (which is based on tf-idf scores) to increase the number of relevant articles for each question. Besides, we also use the \textit{text chunking} and \textit{self-labeled}  techniques introduced in Section~\ref{subs:3} for dealing with long article challenges.

\subsection{Task 5. Statute Law Question Answering}
Task 5 is a newly introduced task in COLIEE 2021.
The goal of this task is to ask systems to perform legal question answering.
In detail, with a given statement, the model needs to answer whether the statement is true or false in the legal aspect.
In essence, Task 5 is constructed from Task 4, ignoring the step of retrieving the related clauses from the Japanese Civil Code.

We face the following challenges in this task: First, the number of training data is relatively small compared to other question answering datasets. Second, the language of the law is different from the daily language. Legal documents are even hard for the lay reader to read. And our mission is to train a deep learning model, a machine to understand the text and give the correct answer.  Dealing with these challenges, first,  we propose to use a pretrained model. This kind of model is pretrained with a massive amount of data and be able to have high performance with less fine-tuning data. Second, there needs to be a method to model the legal language better.

\paragraph{\textbf{Disambiguation Approach}}  The eternal problem of natural language processing in general and legal document processing, in particular, is ambiguity.
The ambiguity can be in the lexical units of the text or in the way a model represents its semantics.
For example, in a text that says ``he went to the bank'', we cannot be certain that the word ``bank'' refers to a river's bank or a financial bank.
This ambiguity can affect the entire inference chain and may lead to serious misunderstandings.

\begin{figure}[htbp]
  \centering
  \includegraphics[width=.5\columnwidth]{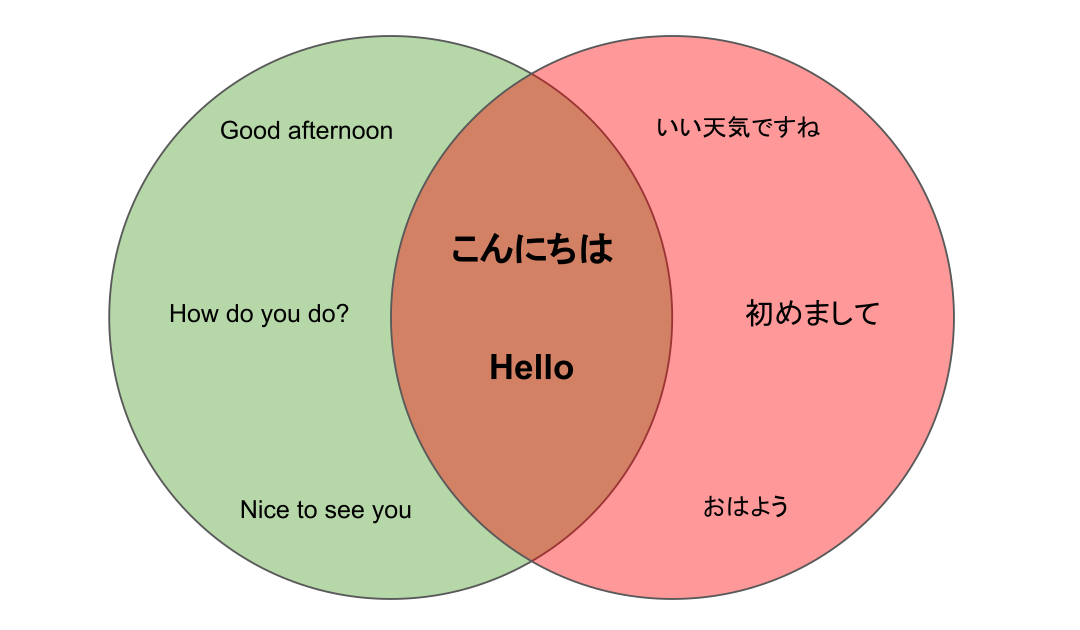}
  
  \caption{A single word may have multiple translations.}
  \label{fig:translation}
\end{figure}

A translation of a text gives us more information about its meaning than just a set of vocabulary translated into a new language.
A sentence in a language may contain much different semantics and depending on the context, the translation needs to be the most appropriate sentence in the target language with the same meaning.
For example, as in Figure~\ref{fig:translation} in Japanese, \begin{CJK*}{UTF8}{min}こんにちは\end{CJK*} can be a midday greeting or a formal way to say ``hello''. In consequence, in the morning context, this sentence needs to be translated as ``hello'' rather than ``good afternoon''.
Similarly, the two meanings of ``bank'' are translated into Japanese with two completely different words, \begin{CJK*}{UTF8}{min}銀行\end{CJK*} and \begin{CJK*}{UTF8}{min}河岸\end{CJK*}.
Likewise, ``I'' in English can be translated in a multitude of ways in Japanese. 
Determining which is the correct translation must depend on the context of the sentence.

Our novelty in Task 5's solution is the introduction of two models, which are NMSP and NFSP.
The main idea in building these two models is to use translation information as means of ambiguity reduction.
We argue that a sentence in natural language can have many meanings, but in its translation, the most correct meaning will be expressed.
In addition, the meaning is also determined by the context, that is, the sentences before and after the current sentence.

It is important to determine the correct context to correctly understand the meaning of a sentence when dealing with difficult documents such as the law.
A correct understanding of semantic will not depend on its language of expression.
Therefore, using the original version and the translation in parallel can help the model learn the semantic with better precision.

\paragraph{\textbf{Pretraining ParaLaw Nets}}  We introduce the pretrained models named ParaLaw Nets~\cite{nguyen2021paralaw}, which are pretrained on cross-lingual sentence-level tasks before being finetuned for use in the COLIEE problem.
The data we use to pretrain these models is bilingual Japanese law data provided by Japanese Law Translation website~\footnote{https://www.japaneselawtranslation.go.jp}.
We formulate the pretraining task for NFSP as a binary classification problem and NMSP as a multi-label classification problem.

We design the pretraining task to force the model to learn the semantic relationship in 2 continuous sentences crossing two languages.
From original sentences as ``The weather is nice. Shall we go out?'', their translations \begin{CJK*}{UTF8}{min}``いい天気ね。お出掛けしよ？''\end{CJK*}, and random sentences as ``Random sentence.'', \begin{CJK*}{UTF8}{min}``ランダム文。''\end{CJK*}, we can generate the training samples as in Table~\ref{tab:pretraining_samples}. 
Following this pattern, we obtain 239,000 samples for pretraining NFSP, and 718,000 samples for pretraining NMSP.

\begin{table}
 \centering
\begin{tabular}{lcc}
\hline
\textbf{Sentence Pair} & \textbf{NFSP Label} & \textbf{NMSP Label} \\
\hline
Shall we go out? The weather is nice.       & -                 & 2  \\
\begin{CJK*}{UTF8}{min}お出掛けしよ？いい天気ね。\end{CJK*}      & -                 & 2  \\
\begin{CJK*}{UTF8}{min}お出掛けしよ？\end{CJK*} The weather is nice.       & -                 & 2  \\
Shall we go out? \begin{CJK*}{UTF8}{min}いい天気ね。\end{CJK*}      & -                 & 2  \\
\hline
\begin{CJK*}{UTF8}{min}いい天気ね。お出掛けしよ？\end{CJK*}      & -                 & 1  \\
The weather is nice. Shall we go out?      & -                 & 1  \\
The weather is nice. \begin{CJK*}{UTF8}{min}お出掛けしよ？\end{CJK*}      & 1                 & 1  \\
\begin{CJK*}{UTF8}{min}いい天気ね。\end{CJK*}Shall we go out?      & 1                 & 1  \\
\hline
The weather is nice. \begin{CJK*}{UTF8}{min}ランダム文。\end{CJK*}      & 0                 & 0  \\
\begin{CJK*}{UTF8}{min}いい天気ね。\end{CJK*}Random Sentence.      & 0                 & 0  \\
The weather is nice. Random Sentence.      & -                 & 0 \\
\begin{CJK*}{UTF8}{min}いい天気ね。ランダム文。\end{CJK*}      & -                 & 0  \\
\hline
\end{tabular}
\caption{Examples of pretraining data.\label{tab:pretraining_samples}}

\end{table}

These models are pretrained until the performance on the validation set does not increase.
Through the results in Table \ref{tab:pretraining_paramters}, we see that the models have better performance on the validation set with NFSP task, base models outperform distilled models. With these numbers, we can accept the assumption that the NFSP task is more straightforward than the NMSP task.

\begin{table}
\small
\centering
\begin{tabular}{lcccc}
\hline
\textbf{Model} & \textbf{Max Len.} & \textbf{Batch Size} & \textbf{\#Batches} & \textbf{Acc.} \\
\hline
NFSP Base       & 512                 & 16                  & 24,000                     & 94.4\%                       \\
NFSP Distilled  & 512                 & 32                  & 34,000                     & 92.2\%                       \\
NMSP Base       & 512                 & 16                  & 320,000                    & 88.0\%                       \\
NMSP Distilled  & 512                 & 32                  & 496,000                    & 87.7\%                       \\
\hline
\end{tabular}
\caption{Parameters and performances in pretraining the models on the validation set.}
\label{tab:pretraining_paramters}
\end{table}

\paragraph{\textbf{Finetuning}} 
Next, we finetune the models for the lawfulness classification problem in Task 5 of the COLIEE-2021. Given a statement as a legal question, the model needs to decide whether that statement is true or false.
Without the support of lexical-based retrieval systems, the model needs to really understand the meaning of the previously learned propositions, generalize them and apply that knowledge to the question. For example, with the sentence ``No abuse of rights is permitted.'' the model should output a ``Yes'' prediction and the predicted label should be ``No'' for the sentence ``The age of majority is reached when a person has reached the age of 12.''

To strengthen the bilingual model, we use original and augmented data in both English and Japanese.
Negation is the main method used to create variations of original data.
The first negation rule that is matched will be used only once to avoid the negation of the negation.
With English negation rules, we reuse the rules proposed by Nguyen et al~\cite{jnlp_task4_coliee2019}.
Japanese negation rules are derived from basic Japanese syntax.
There are in total 14 English negation rules and 13 Japanese negation rules in our rule set.

\section{Experiments}
\subsection{Task 1 and Task 2. Case Law Processing}
As we mentioned in Section~\ref{sec:method}, we create a supporting training dataset to train BERT model and the analysis of this training dataset is as Table \ref{tab:supporting_dataset_analysis}. From 4,415 cases in Task~1 raw dataset, we can extract more than 170K paragraphs. However, inside these paragraphs, some of them contain a lot of french text. This actively demonstrates that we need a filter step to extract clean English text. For the purpose of avoiding noise in the dataset, we use langdetect\footnote{https://pypi.org/project/langdetect/} to filter and remove french text from training data. The clean paragraphs are obtained so far, we split every single paragraph into sentences (>625K in total), and from these sentences, we apply some techniques to generate supporting examples. The massive number of silver training examples for training BERT is over 350K examples.

\begin{table}
\small
\centering
\begin{tabular}{crrrr}

\toprule
\textbf{Data Source}    & \textbf{\#case} & \textbf{\#paragraph} & \textbf{\#sentence} & \textbf{\#example} \\
\toprule
Task 1          & 4,415    & 172,495    & 626,540 &       378,720             \\
Task 2     & 425 & - & 913  &      18,238          \\
\bottomrule
\end{tabular}
\caption{Supporting dataset.}
\label{tab:supporting_dataset_analysis}
\end{table}

Besides silver data, we utilize the data from Task~2 to extract more training data. As you can see in the Table \ref{tab:supporting_dataset_analysis}, the number of gold examples we can extract is more than 18K.

For Task~1, we submit 3 runs as follow:
\begin{itemize}
    \item Run 1 \textit{(BM25SD\_7\_3)}: The lexical score combine with the semantic score with ratio $\alpha$ 7:3.
    \item Run 2 \textit{(BM25SD\_3\_7)}: The lexical score combine with the semantic score with ratio $\alpha$ 3:7.
    \item Run 3 \textit{(SD)}: Only supporting score.
\end{itemize}

For Task~2, we submit 3 runs as follow:

\begin{itemize}
    \item Run1 \textit{(BM25Supporting\_Denoising)}: The lexical score combine with the semantic score with ratio $\alpha$ 7:3.
    \item Run2 \textit{(BM25Supporting\_Denoising\_Finetune)}: The lexical score combine with the semantic score with ratio $\alpha$ 7:3 and finetuning on gold training dataset.
    \item Run3 \textit{(NFSP\_BM25)}: The lexical score combine with NFSP model's score.
\end{itemize}

\begin{table}
\centering
\begin{tabular}{lr}
\hline
 \textbf{Run ID}                     & \textbf{F1 Score}     \\ \hline
BM25SD\_3\_7            & 0.0019               \\
 BM25SD\_7\_3   & 0.0019                \\ 
 SD                           & 0.0019               \\ \hline

\end{tabular}
\caption{Result on Task~1.}
\label{tab:Task1_result}
\end{table}

\begin{table}
\centering
\begin{tabular}{lr}
\hline
 \textbf{Run ID}                     & \textbf{F1 Score}     \\ \hline
BM25Supporting\_Denoising            & 0.6116               \\
 BM25Supporting\_Denoising\_Finetune   & 0.6091                \\ 
 NFSP\_BM25                           & 0.5868               \\ \hline

\end{tabular}
\caption{Result on Task~2.}
\label{tab:Task2_result}
\end{table}

Table \ref{tab:Task1_result} and Table \ref{tab:Task2_result} show our final runs' performance. Our method on Task~1 has bad performance on the test set.
This may indicate that our method has not solved the problem of the large search space of 4,415 candidates for this year's problem.
When we limit the searching space from 4,415 to 100 using the lexical matching score, the actual relevant cases have been filtered out.
Even so, this finding is useful for our future approaches in the next year's competition.
Because of a smaller search space, this problem does not occur with Task 2 data.
For Task~2, our best run achieves 61\% on F1 score, and we are in 5th place in the COLIEE 2021's leaderboard.

\subsection{Task 3. The Statute Law Retrieval Task}
We leverage the last year's dataset to train and evaluate our proposed models. In our experiments, we report the macro- precisions, recals, and $F_{2}$ scores of class-wise precision means and class-wise recall means.

Multiple experiments with different settings are conducted to find the final settings for the text chunking technique. We finetuned on \textit{bert-base-japanese}\footnote{https://huggingface.co/cl-tohoku/bert-base-japanese} pretrained model with 3 epochs and report results in Table~\ref{tab:task3_results_chunk}. Besides, we also did experiments with other pretrained models, e.g., \textit{bert-base-japanese-whole-word-masking}\footnote{https://huggingface.co/cl-tohoku/bert-base-japanese-whole-word-masking} and \textit{xlm-roberta-base}\footnote{https://huggingface.co/xlm-roberta-base}. We use <$window\_size$>/<$stride$> to denote the sliding window parameters. As we can see, the <$150$>/<$50$> setting seems to be the most appropriate setting for this task. We utilize this setting to conduct experiments relating to the self-training technique. In detail, we firstly finetuned \textit{bert-base-japanese} model with $e_{1}$ epochs, then performed the self-labeling process, and continued to finetune with $e_{2}$ epochs. We use <$e_{1}$>/<$e_{2}$> to denote this settings. As we can see in the Table~\ref{tab:task3_results_self_labeled}, it suggests that $e_{1} = 2$ seems to be the “just right” parameter for the first finetuning process (when $e_{1} = 3$, the finetuned model seems to start overfitting); and as $e_{2}$ increases, $F_{2}$ tends to increase. As an observation, our proposed methods demonstrate a positive impact where $F2$ score improve 1.04\% to 72.66\% when applying the text chunking technique, and this number is 1.29\% with the self-labeled technique.

In the competition, we submitted 3 runs based on the proposed approach. For task 3, our position is the runner-up. Table~\ref{tab:task3_runs_results} shows the results of final runs of all participants.

\begin{table}
\small
\centering
\begin{tabular}{lccrrr}
\hline
\textbf{Chunking info.}&\textbf{Return}&\textbf{Retrieved}&\textbf{P}&\textbf{R}&\textbf{F2} \\
\hline
no chunking       &  118   &    81     &       68.24       &     72.52      &   71.62   \\
110/20 &  190   &  {\ul108}  &       61.20       &     67.87      &   66.42   \\
150/10 &  122   &    85     &       64.77       &     66.67      &   66.28   \\
150/20 &  129   &    93     &       68.09       &     70.72      &   70.18   \\
150/40 &  131   &    92     &       67.12       &     71.17      &   70.32   \\
150/50 &  132   &    94   &     {\ul69.74}     &   {\ul73.42}    & {\ul72.66}\\
200/50 &  139   &    90     &       67.12       &     72.97      &   71.72   \\
300/50 &  110   &    74     &       65.39       &     67.57      &   67.12   \\
\hline
\end{tabular}
\caption{(Task 3) Results of \textit{bert-base-japanese} pretrained model finetuning with the text chunking technique. The  values in the \textit{Chunking info.} column denotes chunking settings with format $\langle window\_size\rangle$/$\langle stride\rangle$.}
\label{tab:task3_results_chunk}
\end{table}

\begin{table}
\centering
\begin{tabular}{lccrrr}
\hline
\textbf{Setting}&\textbf{Return}&\textbf{Retrieved}&\textbf{P}&\textbf{R}&\textbf{F2} \\
\hline
3/0&132&94&\underline{69.74}&73.42&72.66\\
1/1&\underline{109}&81&68.02&67.57&67.66\\
1/2&145&98&66.89&72.52&71.32\\
1/3&133&96&68.77&72.97&72.09\\
2/1&161&95&64.55&72.52&70.77\\
2/2&161&95&64.55&72.52&70.77\\
2/3&195&\underline{104}&62.39&\underline{76.13}&\underline{72.91}\\
3/1&146&97&63.32&68.92&67.72\\
3/2&146&96&64.37&71.17&69.70\\
3/3&147&97&60.02&65.77&64.53\\
\hline
\end{tabular}
\caption{(Task 3) Results of \textit{bert-base-japanese} pretrained model with the self-labeled technique.  The  values in the \textit{Setting} column follows the format <$e_{1}$>/<$e_{2}$>.}
\label{tab:task3_results_self_labeled}
\end{table}

\begin{table}
\small
\centering
\begin{tabular}{lccccc}
\hline
\textbf{Run ID}&\textbf{sid}&\textbf{F2}&\textbf{Prec}&\textbf{Recall}\\
\hline
OvGU\_run1&E/J&0.7302&0.6749&0.7778\\
\underline{JNLP.CrossLMultiLThreshold}&E/J&0.7227&0.6000&0.8025\\
BM25.UA&E/J&0.7092&0.7531&0.7037\\
\underline{JNLP.CrossLBertJP}&E/J&0.7090&0.6241&0.7716\\
R3.LLNTU&E/J&0.7047&0.6656&0.7438\\
R2.LLNTU&E/J&0.7039&0.6770&0.7315\\
R1.LLNTU&E/J&0.6875&0.6368&0.7315\\
\underline{JNLP.CrossLBertJPC15030C15050}&E/J&0.6838&0.5535&0.7778\\
OvGU\_run2&E/J&0.6717&0.4857&0.8025\\
TFIDF.UA&E/J&0.6571&0.6790&0.6543\\
LM.UA&E/J&0.5460&0.5679&0.5432\\
TR\_HB&E/J&0.5226&0.3333&0.6173\\
HUKB-3&J&0.5224&0.2901&0.6975\\
HUKB-1&J&0.4732&0.2397&0.6543\\
TR\_AV1&E/J&0.3599&0.2622&0.5123\\
TR\_AV2&E/J&0.3369&0.1490&0.5556\\
HUKB-2&J&0.3258&0.3272&0.3272\\
OvGU\_run3&E/J&0.3016&0.1570&0.7006\\
\hline
\end{tabular}
\caption{(Task 3) Result of final runs on the test set, underlined run IDs refer to our models.}
\label{tab:task3_runs_results}
\end{table}

\subsection{Task 4. The Legal Textual Entailment Task}
Similar to Task 3,  we also trained and evaluated our proposed models with the dataset in the previous year with the questions having id \textit{R-01-*} as the development set. Because of the relatively small training data, we run 5 times with each setting and report the mean and standard deviation values. For this task, most of the hyperparameters follows settings in \cite{nguyen2020jnlp} where $batch\_size=16$ and $learning\_rate=1e^{-5}$.

\paragraph{\textbf{Pretrained model and Data augmentation}} Firstly, we conducted the experiments to find the most suitable pretrained model for this task, and the most suitable setting for the tf-idf-based data augmentation method  (Table~\ref{tab:t4_multi_model}). Based on the experimental results, we found that the \textit{bert-base-japanese-whole-word-masking} pretrained model is more suitable for this task than others. Besides, the tf-idf-based augmentation data method also help increase the model performance.

\begin{table}
\small
\centering
\begin{tabular}{lccccc }
\hline
\textbf{Model}&\textbf{Origin}& \textbf{tf-idf1} & \textbf{tf-idf2} & \textbf{tf-idf5} & \textbf{tf-idf20} \\
\hline
BertJp       &  $55.9\pm3.7$   &   -   &      -   &     -       &  -   \\
BertJp2       &  $60.4\pm4.1$   &   $61.6\pm5.0$      &        \textbf{$62.7\pm5.7$}     &     $61.3\pm4.4$       &   $58.4\pm2.9$    \\
\hline
\end{tabular}
\caption{(Task 4) Model accuracies with different tf-idf augmentation settings. The name \textit{BertJp}, \textit{BertJp2} indicates that we used a pre-trained \textit{bert-base-japanese} and \textit{bert-base-japanese-whole-word-masking}, respectively. The name column follows the format \textit{tf-idf<number>} where \textit{<number>} denotes the number of augmented articles appended for each question.}
\label{tab:t4_multi_model}
\end{table}

\paragraph{\textbf{Performance stability}} In addition, we found that the performance of the model with a small epoch is fairly unstable. Therefore, we experimented with a bigger number of training epochs (Table~\ref{tab:t4_epoch}). The experimental results demonstrate that the runs with higher epochs tend to achieve more stable accuracies, but the model can be overfitted if we increase the number of epochs too much. Besides, the augmentation data also helps the model performance to be more stable.

\begin{table}
\small
\centering
\begin{tabular}{cccc }
\hline
\textbf{\# epochs}&  \textbf{tf-idf1} & \textbf{tf-idf2} & \textbf{tf-idf5}  \\
\hline
3       &   $61.6\pm5.0$      &         $62.7\pm5.7$     &     $61.3\pm4.4$      \\
10       &  $61.8\pm3.0$   &              $62.9\pm2.2$      &        \uline{$64.7\pm2.5$}    \\
20      &   -  &   $61.6\pm1.8$      &         $64.0\pm1.3$       \\
\hline
\end{tabular}
\caption{(Task 4) Accuracies of \textit{BertJp2} with different training epochs.}
\label{tab:t4_epoch}
\end{table}

\paragraph{\textbf{Long article challenge.}} Finally, to address the long article challenge, we conducted the experiments using the \textit{text chunking} and the \textit{self-labeled} techniques described in Task 3 (Table~\ref{tab:t4_self_lb}). Although the accuracy of models in this setting did not increase, the variant ranges are smaller. It may be because the \textit{text chunking} and the \textit{self-labeled} techniques help eliminate noises in training data.

\begin{table}
\centering
\begin{tabular}{lcc }
\hline
\textbf{Setting}&   \textbf{tf-idf2} & \textbf{tf-idf5}  \\
\hline
1/10       &           $62.9\pm1.7$     &     $63.1\pm1.9$      \\
2/10       &           \uline{$64.3\pm0.5$ }    &        \uline{$64.3\pm0.5$ }    \\
3/10      &           $62.2\pm1.7$      &         $64.1\pm1.0$       \\
\hline
\end{tabular}
\caption{(Task 4) Accuracies of \textit{BertJp2} using the \textit{self-labeled} technique with different settings on chunking data where \textit{window\_size = 150}, \textit{stride = 50}.  The  values in the \textit{Setting} column denotes <$e_{1}$> / <$e_{2}$>.}
\label{tab:t4_self_lb}
\end{table}

The results on the blind test set are shown in the Table~\ref{tab:t4_test_rc} with the id ``$J\-NLP.Enss5C15050$'' refers to the model BertJp2 using augmentation data tf-idf5; ``$JNLP.Enss5C15050SilverE2E10$'' refers to the model BertJp2 using augmentation data tf-idf5, and  <$e_{1}$>/<$e_{2}$> is 2 /10; ``$JNLP.EnssBest$'' refers to the ensemble of both models. 

\begin{table}
\centering
\begin{tabular}{llcc }
\hline
\textbf{Team}&   \textbf{sid} & \textbf{Correct} & \textbf{Acc.}  \\
\hline
HUKB    &      HUKB-2       &    57 & 0.7037      \\
UA       &         UA\_parser  &       54  & 0.6667\\
\underline{JNLP}      &           JNLP.Enss5C15050  &      51 & 0.6296    \\
\underline{JNLP}     &           JNLP.Enss5C15050SilverE2E10   &        51 & 0.6296      \\
\underline{JNLP}      &          JNLP.EnssBest  &     51 & 0.6296   \\
OVGU &          OVGU\_run3      &      48 & 0.5926     \\
TR &          TR-Ensemble      &      48 & 0.5926     \\
KIS &           KIS1      &        44 & 0.5432        \\
UA &           UA\_1st     &        44 & 0.5432        \\
\hline
\end{tabular}
\caption{(Task 4) Results final runs  on the test set. The underlined lines refer to our submissions.}
\label{tab:t4_test_rc}
\end{table}

\subsection{Task 5. Statute Law Question Answering}
We compare our proposed models together and with other cross-lingual and multilingual baselines such as XLM-RoBERTa \cite{conneau2019unsupervised} and original BERT Multilingual~\cite{devlin2018bert}. 
In the 7,000 augmented samples, we divide the train set and validation set with the ratio of 9:1.
The blind test set is later provided by the competition's organization.

Our experiments show that NFSP Base and NMSP Base achieve the best performance and have stable loss decrease. NFSP Distilled, NMSP Distilled and XLM-RoBERTa fail to learn from the data and their performance equivalent to that of random sampling. BERT Multilingual is in the middle of the ranked list of models in Table~\ref{tab:finetune_result_validation}. Therefore, we choose NFSP Base, NMSP Base and original BERT Multilingual as candidates for the final run.

\begin{table}
\centering
\begin{tabular}{lr}

\toprule
\textbf{Model}    & \textbf{Accuracy} \\
\toprule
NFSP Base          & 71.0\%                       \\
NFSP Distilled     & 51.1\%                       \\
\midrule
NMSP Base          & 79.5\%                       \\
NMSP Distilled     & 48.9\%                       \\
\midrule
XLM-RoBERTa       & 51.1\%                       \\
BERT Multilingual & 64.1\%                      \\
\bottomrule
\end{tabular}
\caption{(Task 5) Performance of models on the validation set.}
\label{tab:finetune_result_validation}
\end{table}

Table~\ref{tab:final_runs} is the result of the models on the blind test of COLIEE-2021's organizer. On this test set, NFSP Base outperforms other methods and becomes the best system for this task. NMSP is in third place and the original BERT Multilingual has the performance below the baseline. These results again support our proposal in pretraining models using sentence-level cross-lingual information.

\begin{table}
\small
\centering
\begin{tabular}{lllr}
\hline
\textbf{Team} & \textbf{Run ID}                     & \textbf{Correct} & \textbf{Accuracy}     \\ \hline
\underline{JNLP}    & \underline{JNLP.NFSP}                     & \underline{49}         & \underline{\textbf{0.6049}} \\ 
UA            & UA\_parser                          & 46               & 0.5679                \\ 
\underline{JNLP}    & \underline{JNLP.NMSP}                     & \underline{45}         & \underline{0.5556}          \\ 
UA            & UA\_dl                              & 45               & 0.5556                \\ 
TR            & TRDistillRoberta                    & 44               & 0.5432                \\
KIS           & KIS\_2                              & 41               & 0.5062                \\ 
KIS           & KIS\_3                              & 41               & 0.5062                \\ 
UA            & UA\_elmo                            & 40               & 0.4938                \\ 
\underline{JNLP}    & \underline{JNLP.BERT\_Multilingual} & \underline{38}         & \underline{0.4691}          \\ 
KIS           & KIS\_1                              & 35               & 0.4321                \\ 
TR            & TRGPT3Ada                           & 35               & 0.4321                \\ 
TR            & TRGPT3Davinci                       & 35               & 0.4321                \\ \hline
\end{tabular}

\caption{(Task 5) Result of final runs on the test set, the underlined lines refer to our models.}
\label{tab:final_runs}
\end{table}

\section{Conclusions}
This paper presents the approaches of JNLP team at COLIEE 2021 on all 5 tasks of legal document processing. 
The common point of the proposed approaches is the implementation of the Transformer architecture and its variants for specific legal problems. 
The article is not only a technical report on using Transformer for legal processing tasks but also provides details on the methods such as problem definition, data processing, model pretraining and finetuning. 
These proposed methods all come from a detailed investigation of the properties of the problem and data provided in the competition. 
Besides, the NFSP and NMSP construction methods proposed in the paper are novel and effective.
They deserve to be investigated in problems of similar characteristics.




\section{Acknowledgements}
This work was supported by JSPS Kakenhi Grant Number 20H04295, 20K20406, and 20K20625. The research also was supported in part by the Asian Office of Aerospace R\&D (AOARD), Air Force Office of Scientific Research (Grant no. FA2386-19-1-4041).

\bibliographystyle{unsrt}  
\bibliography{references}

\begin{thebibliography}{10}

\bibitem{vaswani2017attention}
Ashish Vaswani, Noam Shazeer, Niki Parmar, Jakob Uszkoreit, Llion Jones,
  Aidan~N Gomez, Lukasz Kaiser, and Illia Polosukhin.
\newblock Attention is all you need.
\newblock {\em arXiv preprint arXiv:1706.03762}, 2017.

\bibitem{devlin2018bert}
Jacob Devlin, Ming-Wei Chang, Kenton Lee, and Kristina Toutanova.
\newblock Bert: Pre-training of deep bidirectional transformers for language
  understanding.
\newblock {\em arXiv preprint arXiv:1810.04805}, 2018.

\bibitem{lan2019albert}
Zhenzhong Lan, Mingda Chen, Sebastian Goodman, Kevin Gimpel, Piyush Sharma, and
  Radu Soricut.
\newblock Albert: A lite bert for self-supervised learning of language
  representations.
\newblock {\em arXiv preprint arXiv:1909.11942}, 2019.

\bibitem{clark2020electra}
Kevin Clark, Minh-Thang Luong, Quoc~V Le, and Christopher~D Manning.
\newblock Electra: Pre-training text encoders as discriminators rather than
  generators.
\newblock {\em arXiv preprint arXiv:2003.10555}, 2020.

\bibitem{lewis2019bart}
Mike Lewis, Yinhan Liu, Naman Goyal, Marjan Ghazvininejad, Abdelrahman Mohamed,
  Omer Levy, Ves Stoyanov, and Luke Zettlemoyer.
\newblock Bart: Denoising sequence-to-sequence pre-training for natural
  language generation, translation, and comprehension.
\newblock {\em arXiv preprint arXiv:1910.13461}, 2019.

\bibitem{radford2018improving}
Alec Radford, Karthik Narasimhan, Tim Salimans, and Ilya Sutskever.
\newblock Improving language understanding by generative pre-training, 2018.

\bibitem{radford2019language}
Alec Radford, Jeffrey Wu, Rewon Child, David Luan, Dario Amodei, and Ilya
  Sutskever.
\newblock Language models are unsupervised multitask learners.
\newblock {\em OpenAI blog}, 1(8):9, 2019.

\bibitem{brown2020language}
Tom~B Brown, Benjamin Mann, Nick Ryder, Melanie Subbiah, Jared Kaplan, Prafulla
  Dhariwal, Arvind Neelakantan, Pranav Shyam, Girish Sastry, Amanda Askell,
  et~al.
\newblock Language models are few-shot learners.
\newblock {\em arXiv preprint arXiv:2005.14165}, 2020.

\bibitem{kowalski2021logical}
Robert Kowalski and Akber Datoo.
\newblock Logical english meets legal english for swaps and derivatives.
\newblock {\em Artificial Intelligence and Law}, pages 1--35, 2021.

\bibitem{satoh2010proleg}
Ken Satoh, Kento Asai, Takamune Kogawa, Masahiro Kubota, Megumi Nakamura,
  Yoshiaki Nishigai, Kei Shirakawa, and Chiaki Takano.
\newblock Proleg: an implementation of the presupposed ultimate fact theory of
  japanese civil code by prolog technology.
\newblock In {\em JSAI International Symposium on Artificial Intelligence},
  pages 153--164. Springer, 2010.

\bibitem{cooper1971definition}
William~S Cooper.
\newblock A definition of relevance for information retrieval.
\newblock {\em Information storage and retrieval}, 7(1):19--37, 1971.

\bibitem{luhn1957statistical}
Hans~Peter Luhn.
\newblock A statistical approach to mechanized encoding and searching of
  literary information.
\newblock {\em IBM Journal of research and development}, 1(4):309--317, 1957.

\bibitem{salton1988term}
Gerard Salton and Christopher Buckley.
\newblock Term-weighting approaches in automatic text retrieval.
\newblock {\em Information processing \& management}, 24(5):513--523, 1988.

\bibitem{sugathadasa2018legal}
Keet Sugathadasa, Buddhi Ayesha, Nisansa de~Silva, Amal~Shehan Perera, Vindula
  Jayawardana, Dimuthu Lakmal, and Madhavi Perera.
\newblock Legal document retrieval using document vector embeddings and deep
  learning.
\newblock In {\em Science and Information Conference}, pages 160--175.
  Springer, 2018.

\bibitem{kien-etal-2020-answering}
Phi~Manh Kien, Ha-Thanh Nguyen, Ngo~Xuan Bach, Vu~Tran, Minh~Le Nguyen, and
  Tu~Minh Phuong.
\newblock Answering legal questions by learning neural attentive text
  representation.
\newblock In {\em Proceedings of the 28th International Conference on
  Computational Linguistics}, pages 988--998, Barcelona, Spain (Online),
  December 2020. International Committee on Computational Linguistics.

\bibitem{liu2019roberta}
Yinhan Liu, Myle Ott, Naman Goyal, Jingfei Du, Mandar Joshi, Danqi Chen, Omer
  Levy, Mike Lewis, Luke Zettlemoyer, and Veselin Stoyanov.
\newblock Roberta: A robustly optimized bert pretraining approach.
\newblock {\em arXiv preprint arXiv:1907.11692}, 2019.

\bibitem{tran2019building}
Vu~Tran, Minh~Le Nguyen, and Ken Satoh.
\newblock Building legal case retrieval systems with lexical matching and
  summarization using a pre-trained phrase scoring model.
\newblock In {\em Proceedings of the Seventeenth International Conference on
  Artificial Intelligence and Law}, pages 275--282, 2019.

\bibitem{10.1145/3322640.3326741}
Juliano Rabelo, Mi-Young Kim, and Randy Goebel.
\newblock Combining similarity and transformer methods for case law entailment.
\newblock In {\em Proceedings of the Seventeenth International Conference on
  Artificial Intelligence and Law}, ICAIL '19, page 290–296, 2019.

\bibitem{nguyen2020jnlp}
Ha-Thanh Nguyen, Hai-Yen~Thi Vuong, Phuong~Minh Nguyen, Binh~Tran Dang,
  Quan~Minh Bui, Sinh~Trong Vu, Chau~Minh Nguyen, Vu~Tran, Ken Satoh, and
  Minh~Le Nguyen.
\newblock Jnlp team: Deep learning for legal processing in coliee 2020.
\newblock {\em arXiv preprint arXiv:2011.08071}, 2020.

\bibitem{jnlp_3_2019}
T.B.Dang, T.Nguyen, and L.M.Nguyen.
\newblock An approach to statute law retrieval task in coliee-2019., 2019.

\bibitem{dbse_3_2019}
S.Wehnert, S.A.Hoque, W.Fenske, and G.Saake.
\newblock Threshold-based retrieval and textual entailment detection on legal
  bar exam questions., 2019.

\bibitem{iitp_3_2019}
B.Gain, D.Bandyopadhyay, T.Saikh, and A.Ekbal.
\newblock Iitp@coliee 2019: Legal information retrieval using bm25 and bert.,
  2019.

\bibitem{kis_3_2019}
R.Hayashi and Y.Kano.
\newblock Searching relevant articles for legal bar exam by doc2vec and
  tf-idf., 2019.

\bibitem{kis_4_2019}
R.Hoshino, N.Kiyota, and Y.Kano.
\newblock Question answering system for legal bar examination using predicate
  argument structures focusing on exceptions., 2019.

\bibitem{tr_4_2019}
J.Hudzina, T.Vacek, K.Madan, C.Tonya, and F.Schilder.
\newblock Statutory entailment using similarity features and decomposable
  attention models., 2019.

\bibitem{jnlp_task4_coliee2019}
HT~Nguyen, V~Tran, and LM~Nguyen.
\newblock A deep learning approach for statute law entailment task in
  coliee-2019.
\newblock {\em Proceedings of the 6th Competition on Legal Information
  Extraction/Entailment. COLIEE}, 2019.

\bibitem{triguero2015self}
Isaac Triguero, Salvador Garc{\'\i}a, and Francisco Herrera.
\newblock Self-labeled techniques for semi-supervised learning: taxonomy,
  software and empirical study.
\newblock {\em Knowledge and Information systems}, 42(2):245--284, 2015.

\bibitem{nguyen2021paralaw}
Ha-Thanh Nguyen, Vu~Tran, Phuong~Minh Nguyen, Thi-Hai-Yen Vuong, Quan~Minh Bui,
  Chau~Minh Nguyen, Binh~Tran Dang, Minh~Le Nguyen, and Ken Satoh.
\newblock Paralaw nets--cross-lingual sentence-level pretraining for legal text
  processing.
\newblock {\em arXiv preprint arXiv:2106.13403}, 2021.

\bibitem{conneau2019unsupervised}
Alexis Conneau, Kartikay Khandelwal, Naman Goyal, Vishrav Chaudhary, Guillaume
  Wenzek, Francisco Guzm{\'a}n, Edouard Grave, Myle Ott, Luke Zettlemoyer, and
  Veselin Stoyanov.
\newblock Unsupervised cross-lingual representation learning at scale.
\newblock {\em arXiv preprint arXiv:1911.02116}, 2019.

\end{thebibliography}

\end{document}